%% file: main.tex
\documentclass[10pt,twocolumn,letterpaper]{article}

\usepackage[pagenumbers]{cvpr} 

\input{preamble}
\definecolor{cvprblue}{rgb}{0.21,0.49,0.74}
\usepackage[pagebackref,breaklinks,colorlinks,allcolors=cvprblue]{hyperref}
\usepackage{adjustbox}
\usepackage{multirow}
\usepackage{multicol}
\usepackage{makecell}
\usepackage{colortbl}
\usepackage{nicefrac} 
\usepackage{bm}
\usepackage{booktabs}
\usepackage{amsthm}
\usepackage{pifont}

\newtheorem{Thm}{Theorem}

\def\paperID{xxxxx} 
\def\confName{CVPR}
\def\confYear{2026}

\title{Vanilla Group Equivariant Vision Transformer: Simple and Effective}

\author{Jiahong Fu$^{1}$ \hspace{2em}  Qi Xie$^{1}$\hspace{2em} Deyu Meng$^{1}$\hspace{2em} Zongben Xu$^{1}$\\
$^{1}$  School of Mathematics and Statistics, Xi’an Jiaotong University, Xi'an, China
}

\begin{document}
\maketitle
\input{sec/0_abstract}    
\input{sec/1_intro}
\input{sec/2_related}

\input{sec/3_method}
\input{sec/4_exp}
\input{sec/5_con}
{
    \small
    \bibliographystyle{ieeenat_fullname}
    \bibliography{main}
}


\end{document}

%% file: sec/0_abstract.tex
\begin{abstract}
Incorporating symmetry priors as inductive biases to design equivariant Vision Transformers (ViTs) has emerged as a promising avenue for enhancing their performance. However, existing equivariant ViTs often struggle to balance performance with equivariance, primarily due to the challenge of achieving holistic equivariant modifications across the diverse modules in ViTs—particularly in harmonizing the Self-Attention mechanism with Patch Embedding. To address this, we propose a straightforward framework that systematically renders key ViT components, including patch embedding, self-attention, and positional encodings, and Down/Up-Sampling, equivariant, thereby constructing ViTs with guaranteed equivariance. The resulting architecture serves as a plug-and-play replacement that is both theoretically grounded and practically versatile, scaling seamlessly even to Swin Transformers. Extensive experiments demonstrate that our equivariant ViTs consistently improve performance and data efficiency across a wide spectrum of vision tasks.



\end{abstract}

%% file: sec/1_intro.tex
\section{Introduction}
\label{sec:intro}

Vision Transformers (ViTs) \cite{dosovitskiy2020image, liu2021swin, touvron2021training} have become a dominant architecture in computer vision, achieving state-of-the-art performance across diverse tasks including image classification, detection, segmentation, and restoration. This success stems primarily from the self-attention mechanism, which excels at capturing long-range dependencies and has established ViTs as a foundational paradigm surpassing convolutional and fully-connected networks in modeling global context.

Current research \cite{conwell2024large, dieleman2016exploiting, cohen2016steerable, rojas2024making, ding2023reviving} have shown that proper inductive biases are important for enhancing the performance of deep network architectures, including the Self-Attention mechanism.  Particularly,  in computer vision tasks, transformations like image translation, rotation, and mirroring usually do not affect the final image recognition and classification results \cite{agrawal2015learning, cohen2016group}, which naturedly attribute to the most common used inductive bias: symmetry priors present  in  both  the local features and global semantics under the aforementioned transformations. However, standard Vision Transformers (ViTs) have not explicitly embedded such inductive biases into their design. Consequently, current ViT models typically rely on massive datasets and extensive data augmentation to implicitly learn these symmetries—a process that demands substantial data and computational resources while still failing to guarantee the systematic embedding of symmetry priors \cite{rojas2024making, ding2023reviving, fan2021multiscale}.


It is noteworthy that in recent years, the emergence of equivariant CNN architectures \cite{cohen2016group, shen2020pdo, xie2022fourier} has achieved remarkable success in embedding symmetry priors, demonstrating that directly designing equivariant model architectures offers unique advantages in parameter efficiency and performance enhancement, compared to learning symmetries from training data.
This insight inspires the recognition that designing  guaranteed  equivariant architecture for ViTs is a highly promising research direction.

However, replicating the successful design of equivariant architectures from CNNs to Vision Transformers (ViT) has been proven to be challenging.
Most existing approaches leverage the intrinsic permutation equivariance of self-attention \cite{romero2020group, xu2023e2}, yet this property is disrupted during patch embedding, forcing methods to operate directly on pixel-level features. This approach significantly compromises local feature representation and generally underperforms modern ViT architectures \cite{dosovitskiy2020image, liu2021swin, touvron2021training}. Furthermore, quadratic memory growth with image size limits processing to low-resolution inputs. Recent theoretical frameworks for ViT equivariance have been proposed \cite{he2021efficient, kundu2024steerable}, but resulting implementations are often overly complex and fail to surpass standard ViT performance. More importantly, these methods differ too significantly from the original ViT, making them difficult to adapt to ViT variants, such as Swin-Transformers \cite{liu2021swin}, which largely restricts their applicability.

The aforementioned methods' struggle in balancing performance and equivariance highlights two fundamental difficulties in designing equivariant ViTs: \textbf{(I) Effective equivariant designs that harmonize the Self-Attention mechanism with Patch Embedding remain elusive.} Unlike the convolution in CNNs, which is inherently equivariant to local translations, Self-Attention possesses long-range connectivity and a more complex computational structure. This complexity makes its equivariant modification significantly more challenging, especially when patches themselves undergo rotations or reflections. \textbf{(II) A holistic equivariant framework for diverse other modules in ViTs is lacking.} Beyond the above two core components, standard Vision Transformers incorporate various other modules, such as positional encodings, Down/Up-Sampling layers(in  Swin-Transformer), LayerNorm, all of which are essential for performance. All of these components must be made equivariant holistically to guarantee the equivariance of the entire architecture.

To address the aforementioned challenges, we explore a vanilla path to construct comprehensive equivariant framework for all common components in Vision Transformers, thereby constructing ViT architectures that are rigorously equivariant to $\nicefrac{\pi}{2}$ rotations and reflections. Our principal contributions are as follows:

(1) We propose a straightforward method to convert a standard Vision Transformer (ViT) into its rotation and reflection equivariant version. First, we employ an equivariant CNN-based patch embedding to lift the input domain, producing feature maps with an added group dimension. We then use equivariant linear layers \cite{ravanbakhsh2020universal, xie2025rotation} to harmonize Self-Attention with the patch embedding in an equivariant manner. Furthermore, via judicious group-wise parameter sharing and a tailored feature reordering strategy, we redesign positional encodings and Resampling layers, forming a holistic equivariant framework for diverse ViT modules. As a result, our equivariant ViT acts as a plug-and-play replacement across applications, enabling potential performance gains and parameter reduction.

(2) For the first time, our method can be seamlessly and effectively applied to Swin Transformer, enhancing its performance. This successful adaptation significantly broadens the applicability and practicality of equivariant architectures, demonstrating our framework's generality.

(3) Through rigorous theoretical analysis, we prove that the proposed framework maintains strict equivariance for both the overall network architecture and individual modules. Simultaneously, our theoretical findings indicate that the modified ViT architecture will possess an enhanced generalization capability.

(4) Extensive experiments across multiple benchmarks validate our method's effectiveness in consistently improving performance and data efficiency across various tasks, from high-level recognition to low-level restoration. These results establish that explicit geometric symmetry embedding is crucial for developing more powerful and robust Vision Transformers.

%% file: sec/2_related.tex
\section{Related Work}
\vspace{-2mm}
\label{sec:related}

\subsection{Vision Transformers}
The Transformer architecture originally developed for NLP~\cite{vaswani2017attention} has become foundational in computer vision. The Vision Transformer (ViT)~\cite{dosovitskiy2020image} first adapted it by treating images as sequences of $16 \times 16$ patches, revealing self-attention's potential in vision tasks. However, ViT suffers from quadratic complexity and limited inductive bias, restricting its scalability and efficiency. The Swin Transformer~\cite{liu2021swin, liu2022swinv2} addressed this with a hierarchical shifted-window design, enabling linear complexity while retaining global context, later extended to restoration tasks in SwinIR~\cite{liang2021swinir}.
Beyond pure attention models, hybrids like CvT~\cite{wu2021cvt} integrate convolutional projections to embed spatial locality efficiently. Other approaches~\cite{wang2021pyramid, fan2021multiscale, li2022mvitv2} use pyramidal structures for multi-scale modeling. Recently, Vision Transformers are increasingly tailored to specific tasks: SwinIR leverages hierarchy for restoration, while DETR~\cite{carion2020end} and SegFormer~\cite{xie2021segformer} adapt architectures for detection and segmentation.

\subsection{Group Equivariant Neural Networks}
Early attempts for exploring transformation symmetry priors in images primarily rely on data augmentation \cite{simard2003best, krizhevsky2012imagenet, laptev2016ti} and self-supervised learning frameworks \cite{agrawal2015learning, dieleman2016exploiting}. Recently, most works have primarily focused on embedding transformation equivariance directly into network architectures via equivariant module designs. Prior work in this direction is G-CNN~\cite{cohen2016group}, it achieves 90-degree rotation and mirror reflection equivariance. Since then various types of rotationally equivariant convolution \cite{shen2020pdo, e2cnn} networks have emerged. In recent years, research on the equivariance of Transformer model embeddings has been quite active. In recent years, there have been many efforts \cite{romero2020group, xu2023e2, kundu2024steerable, bokman2025flopping,he2021efficient} to design equivariant vision transformers. Specifically, \cite{romero2020group, xu2023e2} construct equivariant Self-Attention by a lifting positional encoding. \cite{he2021efficient} constructs an efficient equivariant attention layer and \cite{kundu2024steerable} design a sterrable ViT. Further, shift equivariant Transformer \cite{rojas2024making, ding2023reviving} are beginning to attract attention.



%% file: sec/3_method.tex
\section{Method}

In this section, we first present the foundational concepts for equivariance and vision transformer architecture, and then provide the proposed equivariant ViT, as well as the theoretical analysis for equivariance error and generalization error.

\begin{figure*}[ht]
    \centering
    \includegraphics[width=17cm]{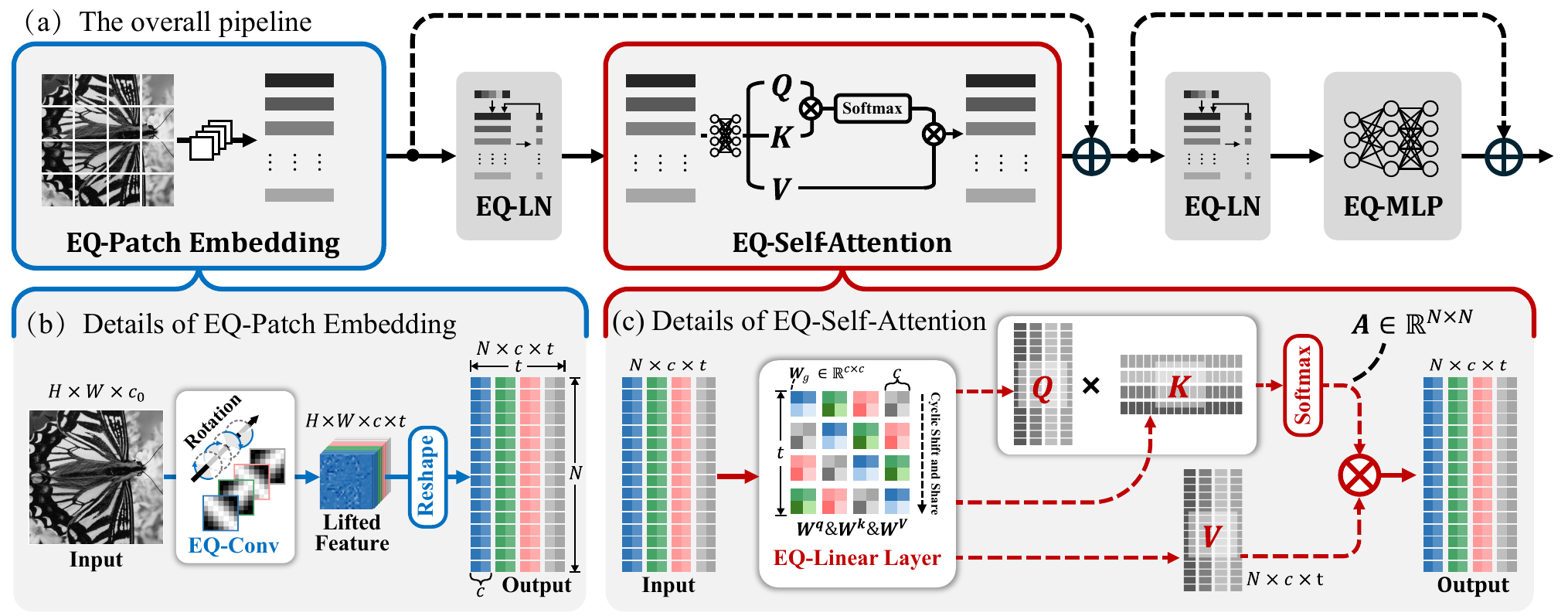}
    \vspace{-2mm}
    \caption{Overview of the proposed Equivariant Vision Transformer. (a) The overall pipeline of the whole equivariant vision transformer. (b)-(c) The details of the EQ-Patch Embedding and the EQ-Self-Attention, respectively.}
    \label{fig:method}
    \vspace{-4mm}
\end{figure*}

\subsection{Preliminaries}
{\bf Equivariance.} Equivariance of a mapping means that a transformation on the input will result in a predictable transformation on the output \cite{cohen2019general, shen2020pdo, xie2022fourier}. In this work, we focus on achieving rotation and mirror reflection equivariance in Vision Transformer. For example, rotation equivariance ensures that rotating the input leads solely to a corresponding rotation of the output, without introducing any additional, unpredictable variations. Mathematically, let $\Psi$ denote a Vison Transformer mapping from the input feature
space to the output feature space, and $S$ be a group of rotation and mirror reflection transformations, \ie,
\begin{equation}\label{eq2}
S=\! \left\{\!g(k,m) \! = \!
  \begin{bmatrix}
   (-1)^m \cos \nicefrac{2\pi k}{t} & -(-1)^m\sin \nicefrac{2\pi k}{t} \\
   \sin \nicefrac{2\pi k}{t} &  \cos \nicefrac{2\pi k}{t}
  \end{bmatrix} \right\},
\end{equation}
where $k = \left\{0,1,\cdots,t\!-\!1\right\}$ and $m\in\{0,1\}$.
\noindent
Then, $\Psi$ is equivariant with respect to $S$, if for any transformation matrix $\tilde{g}\in{S},$
\begin{equation}\label{eq3}
\Psi\left[\pi_{\tilde{g}}^I\right](I)=\pi_{\tilde{g}}^F\left[\Psi\right](I),
\end{equation}

\noindent
where \textit{I} is an input image, $\pi_{\tilde{g}}^I$ and $\pi_{\tilde{g}}^F$ denote how the transformation $\tilde{g}$ acts on input image and output features, respectively, $[\begin{array}{c}{\cdot}\end{array}]$ denotes the composition of functions.

\noindent
{\bf Self-Attention.} We first revist the attention mechanism in Vision Transformer. The Vision Transformer takes a visual token sequence $\bm{z}_{\ell-1} \in \mathbb{R}^{N \times d}$ from the previous layer ($\ell-1 ~th$ layer) as input ($N$ is the token number and $d$ is the hidden dimension), then projects it into the query, key, and value token sequences with three different linear projection layers, denoted as $\mathbf{W}^{\mathbf{q}}$, $\mathbf{W}^{\mathbf{k}}$, $\mathbf{W}^{\mathbf{v}}\in \mathbb{R}^{d \times d}$:
\begin{equation}
    \bm{q} = \bm{z}_{\ell-1} \cdot \mathbf{W}^{\mathbf{q}}, \ \bm{k} = \bm{z}_{\ell-1} \cdot \mathbf{W}^{\mathbf{k}}, \ \bm{v} = \bm{z}_{\ell-1} \cdot \mathbf{W}^{\mathbf{v}}.
\end{equation}
where $\bm{q}, \bm{k}, \bm{v} \in \mathbb{R}^{N \times d}$ and $\cdot$ denotes the matrix multiplication. Then the similarity of each query $\bm{q}$ and key $\bm{k}$ is computed as:
\begin{equation}
    \mathbf{A} = \operatorname{Softmax} \left( \bm{q} \bm{k}^{\top} / \sqrt{d} \right),
\end{equation}
where the attention map $\mathbf{A}$ is an $N\times N$ matrix with values in the range $[0, 1]$, and the sum of each row is normalized to 1. Then the Self-Attention reweights the value sequences $\bm{v}$ according to the attention map $\mathbf{A}$, i.e. $\mathbf{y} = \mathbf{A} \bm{v} \in \mathbb{R}^{N\times d}$. The multi-head self-attention mechanism is a straightforward extension, involving parallel self-attention computations whose outputs are linearly aggregated, and is thus omitted for brevity.

\noindent
{\bf Positional Encoding.} 
Positional encoding in transformers is used to embed information about the position of tokens within a sequence into their input representations. In Vision Transformers, the absolute position encoding is usually set as a learnable matrix $E_{\operatorname{pos}}^{\mathcal{A}} \in \mathbb{R}^{N \times d}$.

In contrast to ViT~\cite{dosovitskiy2020image}, the Swin-Transformer~\cite{liu2021swin} introduces a more dynamic approach by incorporating a relative position bias. Instead of adding position information directly to the input embeddings, the Swin-Transformer~\cite{liu2021swin} injects positional information into the self-attention mechanism. Specifically, a learnable bias term $E_{\operatorname{pos}}^{\mathcal{R}} \in \mathbb{R}^{N\times N}$ is added to the scaled dot-product attention map $\mathbf{A}$, i.e. $\mathbf{A}+E_{\operatorname{pos}}^{\mathcal{R}}$, modifying the similarity calculation between queries and keys. This bias is determined by the pairwise relative distance between the tokens (image patches) within a local window.


\noindent
{\bf Vision Transformers.}
A standard Vision Transformer first tokenizes an input image $\mathbf{x} \in \mathbb{R}^{C \times H \times W}$ into a sequence of embeddings. This is accomplished by a patch embedding layer, typically implemented as a 2D convolution with a kernel size and stride equal to the patch size $s$. This operation effectively projects each non-overlapping image patch into a $d$-dimensional embedding, resulting in a sequence of patch tokens. The resulting sequence $\bm{z}_0 \in \mathbb{R}^{N\times d}$ is then processed through a stack of $L$ transformer blocks:
\begin{equation}
\label{eq:vit_architecture} 
\begin{aligned}
    \bm{z}_{0} &= [\operatorname{Transpose}\left(\operatorname{Reshape}\left( \text{Conv2D}_s(\mathbf{x})\right)\right)], \\
    \bm{z}'_{\ell} &= \text{MSA}(\text{LN}(\mathbf{z}_{\ell-1})) + \bm{z}_{\ell-1}, & \ell=1 \dots L \\
    \bm{z}_{\ell} &= \text{MLP}(\text{LN}(\mathbf{z}'_{\ell})) + \bm{z}'_{\ell}, & \ell=1 \dots L
\end{aligned}
\end{equation}
where $\operatorname{Reshape} (\cdot)$ denotes reshaping the output of the $\text{Conv2D}_s$ layer into a sequence of tokens and $\operatorname{Transpose} (\cdot)$ represents the reversal of dimensions. Each encoder block consists of a multi-head self-attention (MSA) module and a Multilayer Perceptron (MLP). Layer Normalization (LN) is applied before each module, and a residual connection is used after each.

\subsection{Equivariant Patch Embedding/Self-Attention}

In this section, we begin by detailing the core components of our proposed method: Equivariant Patch Embedding (EQ-PE) and Equivariant-Self-Attention (EQ-SA).

{\bf Equivariant Patch Embedding.} In a conventional Vision Transformer, the input image $\bm{x}\in \mathbb{R}^{H \times W \times c_0}$ is typically tokenized by a Patch Embedding layer to get features $\bm{z}_0$. For the proposed equivariant Vision Transformer, we will need $\bm{z}_0$ to be equivariant with respect to input $\bm{x}$. We found that  equivariant convolution with proper stride  \cite{wu2025measuring} can elegantly fulfill this requirement. Mathematically, we perform the following EQ-PE:
\begin{equation}\label{eq:pe}
\left\{
\begin{array}{l}
\hat{\bm{z}}^{g} = D_s\left(\pi_A(\psi) \otimes \bm{x}\right), \forall g\in S, \\
\bm{z}_0 = \operatorname{Reshape}\left( \hat{\bm{z}} \right),
\end{array}
 \right.
\end{equation}
where $\hat{\bm{z}} \in \mathbb{R}^{\frac{H}{s}\times \frac{W}{s}\times c\times t}$ is the output feature of convolution, with $\hat{\bm{z}}^{g} \in \mathbb{R}^{\frac{H}{s}\times \frac{W}{s}\times c}$ denotes its slice tensor indicated by $g$ in the group dimension, $t$ is the element number of the selected transformation group $S$,   $s$ is the stride  of  downsampling $D_s$\footnote{Equivariant down-sampling will be discussed in a subsequent section.} ; $\psi$ is a to be learned convolution kennel;  $\pi_g$ is the transformation with respected to $g$ (e.g., rotation the  with coordinate transformation matrix $g$); $\otimes$ denoted a 2D convolution with input channel of $c_0$ and output channel of $c$; $\operatorname{Reshape}$ reshape a tensor from $\frac{H}{s}\times \frac{W}{s}\times c\times t$ to  $N\times c\times t$, where $N = \frac{HW}{s^2}$. 

The above calculation is illustrated in detail in Fig. \ref{fig:method}(b) for clarity.
It should be noted that the resulting feature $\bm{z}_0$ achieve by \ref{eq:pe} is equivariant with respect to  the input $\bm{x}$. When the input image is transformed, the feature z moves according its corresponding image patch (patch of the receptive field), with only  a predictable cyclic shift  along the group channel due to the rotation or reflection of the image patch, without introducing any other unpredictable changes.

{\bf Equivariant Self-Attention.} We take the Self-Attention layer with in put feature $\bm{z}\in\mathbb{R}^{N\times c \times t}$ as example (It should be noted that in the first EQ-PE stage we have lifted the features with a additional group dimension of size $t$). 

We obtain the query, key, and value by  applying the following an equivariant linear layers~\cite{ravanbakhsh2020universal} to  the input $\bm{z}$:
\begin{equation}
    \begin{aligned}
    \bm{q}^B &= \sum_{g\in S}  \bm{z}^g_0 \cdot \bm{W}_{B^{-1} g}^{\mathbf{q}}, \forall B\in S,\\
    \bm{k}^B &= \sum_{g\in S}  \bm{z}^g_0 \cdot \bm{W}_{B^{-1} g}^{\mathbf{k}}, \forall B\in S,\\
    \bm{v}^B &= \sum_{g\in S}  \bm{z}^g_0 \cdot \bm{W}_{B^{-1} g}^{\mathbf{v}}, \forall B\in S,
\end{aligned}
\vspace{-1mm}
\end{equation}
where $\bm{q},\bm{k},\bm{v} \in {R}^{N\times c\times t}$, with $\bm{q}^B,\bm{k}^B,\bm{v}^B \in \mathbb{R}^{N\times c}$ is the slice matrix indicated by $B$ in the group dimension, respectively;
$\bm{W}^q_g, \bm{W}^k_g,\bm{W}^v_g\in \mathbb{R}^{c\times c}$, $\forall g\in S$  denotes the learnable parameters, and $\bm{z}^g_0\in \mathbb{R}^{N\times c}$ is the slice matrix of $\bm{z}_0$ indicated by $g$ in the group dimension,.

The above calculation regarding $\bm{q}$ is equivalent to constructing a larger matrix  $\bm{W}\in \mathbb{R}^{ct\times ct}$  by tiling $\bm{W}_{B^{-1} g}^{\mathbf{q}}$, and then directly multiplying it with the matrix $\mathbf{Z}\in \mathbb{R}^{N\times ct}$ formed by  $\left[\bm{z}^{B_1}, \bm{z}^{B_2},\cdots, \bm{z}^{B_t}\right]$, as shown in Fig. \ref{fig:method}(c). It can be observed that in this process, the parameters in $\bm{W}$ exhibit clear  cyclic shifting and sharing  patterns. It can be proven that the linear operation performed in this manner is equivariant \cite{ravanbakhsh2020universal}. The same patterns holds for $\bm{k}$ and $\bm{v}$ too.

After obtaining query, key, and value, we compute the attention map $\bm{A}$ and the final output $\bm{z}_1$
\begin{equation}
    \label{eq:equi_sa}
    \begin{aligned}
     \bm{A} = \operatorname{Softmax}\left( \bm{Q}\cdot\bm{K}^{\top} / \sqrt{ct}\right), 
     \bm{z}_1 = \bm{A}\cdot \bm{V},
    \end{aligned}
\end{equation}
where $\bm{K}, \bm{Q}, \bm{V} \in \mathbb{R}^{N\times ct}$ are matrixes reshaped from $\bm{q}, \bm{k}, \bm{v} \in \mathbb{R}^{N\times c \times t}$. 

The above calculation is illustrated in detail in Fig. \ref{fig:method}(c) for clarity.
A similar process can be intuitively deduced from the construction of equivariant Multi-Head Self-Attention.

\subsection{Equivariant Positional Encoding}
Vision Transformers typically incorporate positional encodings to model spatial relationships between image patches. These are commonly implemented in two forms: absolute position encoding, which assigns a unique embedding to each patch, and relative position encoding, which explicitly models pairwise spatial offsets between patches.
\begin{figure}[h]
    \centering
    \setlength{\abovecaptionskip}{0.1cm}
    \includegraphics[width=8.3cm]{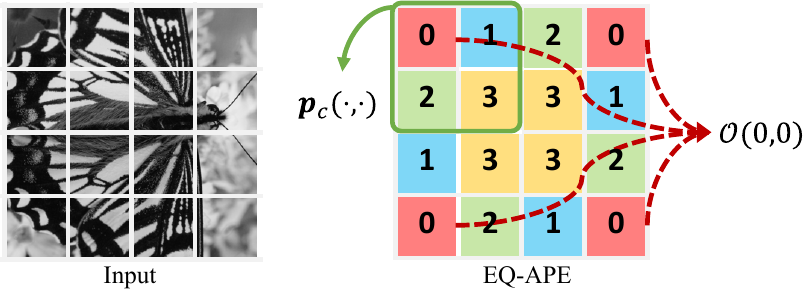}
    \vspace{-5mm}
    \caption{The visualization of the input and the proposed Equivariant APE for $C_4$ group. For EQ-APE, different numbers represent different orbits. And $\mathbf{p}_c(\cdot, \cdot)$ denote the canonical representations}
    \label{fig:ape}
    \vspace{-0.4cm}
\end{figure}

\noindent
{\bf Equivariant Absolute Positional Encoding.}
Let the input image be discretized into a grid of $N \times N$ patches. The position of any patch is indexed by its 2D coordinates $\mathbf{p} \!=\! (i, \!j)$, where $i, j \in \{0, \!\dots, \!N\!-\!1\}$. Let's take the cyclic group $C_4$ of 90-degree rotations as an example \footnote{The set of 90-degree rotations is known in mathematics as the cyclic group C4, contains four distinct rotation orientations: $0^{\circ}$, $90^{\circ}$, $180^{\circ}$, and $270^{\circ}$.}. The action of this group on a position $\mathbf{p}$ generates a set of equivalent positions, known as the orbit $\mathcal{O}(\mathbf{p})$ \footnote{The orbit of an element $x$ under the action of a group $S$ is the set of all elements that can be reached by applying the transformations of $S$ to $x$.}. For any position $(i, j)$:
\begin{equation}
    \begin{aligned}
    \mathcal{O}(i, \!j) \!\!=\!\! 
    \{\!(i, j),\! (j, N\!\!-\!\!1\!\!-\!\!i),\! (N\!\!-\!\!1\!\!-\!i, N\!\!-\!\!1\!\!-\!\!j),\! (N\!\!-\!\!1\!\!-\!\!j, i)\!\},
    \end{aligned}
\end{equation}
and in Figure \ref{fig:ape}, we visualize the orbit of the position $(0,0)$.

To ensure all positions within an orbit share an identical embedding, we first map each position $\mathbf{p}$ to a canonical representative $\mathbf{p}_c$ of its orbit. Specifically, we define the canonical representation, shown in Figure \ref{fig:ape}, as the lexicographically smallest element within the orbit:
\begin{equation}
    \mathbf{p}_c(i, j) = \min_{\text{lex}} \mathcal{O}(i, j),
\end{equation}
where $\min_{\text{lex}}$ denotes the selection of the element that is smallest in lexicographical order. For two-dimensional coordinates, a pair $(i_1, j_1)$ is considered lexicographically smaller than $(i_2, j_2)$ if $i_1 < i_2$, or if $i_1 = i_2$ and $j_1 < j_2$.

For a position $p\!=\!(i,\!j)$ with its orbit representative $p_c(i,j)$, given that the absolute positional encoding $\text{APE}(\cdot)$ is assigned to $p_c$, then all elements within the group orbit of this coordinate share the same APE. Specifically, our $C_4$-invariant absolute position encoding can be formally expressed as:
\begin{equation}
    \text{PE}_{C_4}(i, j) = \text{APE}(\mathbf{p}_c(i, j)).
\end{equation}

This construction guarantees that for any rotation $g \in C_4$, the encoding remains invariant, i.e., $\text{PE}_{C_4}(g(\mathbf{p})) = \text{PE}_{C_4}(\mathbf{p})$, thereby providing the model with a consistent representation of spatial location regardless of the input's orientation. For the construction of the APE for the mirror transformation group, see the supplementary materials.

\noindent
{\bf Equivariant Relative Position Encoding.}
A core objective of relative position encoding (RPE) is to model intrinsic spatial relationships between token pairs, independent of their global pose. However, standard RPEs, typically based on displacement vectors, are sensitive to rotations and reflections, and fail to recognize that transformed pairs remain geometrically congruent.
To address this, we introduce an RPE formulation based on the dihedral group $D_4$, which captures the full set of eight square symmetries: four rotations (0\textdegree, 90\textdegree, 180\textdegree, 270\textdegree) and four reflections (horizontal, vertical, and two diagonals). Our method ensures invariance under $D_4$ by assigning identical positional encodings to any two patch pairs that are equivalent under these transformations.


To achieve this, we generalize the concept of canonical representation to the group $D_4$. We define the orbit of the pair $(\mathbf{p}_i, \mathbf{p}_j)$ as the set of all pairs reachable by applying the same group element $g \in D_4$ to both points:
\begin{equation}
    \mathcal{O}_{D_4}(\mathbf{p}_i, \mathbf{p}_j) = \{ (g(\mathbf{p}_i), g(\mathbf{p}_j)) \mid g \in D_4 \}.
\end{equation}
Within this orbit, which contains all geometrically equivalent configurations of the point pair, we select a \textbf{canonical representative}. This representative, denoted $(\mathbf{p}_{ic}, \mathbf{p}_{jc})$, is chosen as the lexicographically smallest element in the orbit, where the comparison is performed on the flattened 4-tuple representation $(i_1, j_1, i_2, j_2)$:
\begin{equation}
    (\mathbf{p}_{ic}, \mathbf{p}_{jc}) = \min_{\text{lex}} \mathcal{O}_{D_4}(\mathbf{p}_i, \mathbf{p}_j).
\end{equation}

Finally, our $D_4$-invariant relative position encoding is defined by applying a base RPE function (such as a learnable embedding) to the displacement vector of this canonical pair:
\begin{equation}
    \text{PE}_{D_4}^{\text{rel}}(\mathbf{p}_i, \mathbf{p}_j) = \text{RPE}(\mathbf{p}_{jc} - \mathbf{p}_{ic}).
\end{equation}

This construction maps any point pair along with its rotated or reflected counterparts to the same canonical displacement vector, effectively normalizing relative orientation prior to encoding. As a result, the self-attention mechanism gains true geometric invariance, enabling the model to focus on fundamental spatial relationships and improving both generalization capability and data efficiency.

\subsection{Equivariant Down/Up-Sampling}
To capture multi-scale visual features, recent Vision Transformers often employ down-sampling at various stages. This includes initial patch embedding via strided convolution, as well as subsequent down-sampling layers like patch merging, to reduce spatial resolution. Meanwhile, certain tasks utilize up-sampling layers, such as Pixel-Shuffle, to increase resolution. However, these conventional down-sampling and up-sampling operations inherently break group equivariance. Therefore, designing equivariant resampling layers is a necessary step toward building fully equivariant Vision Transformers.

\noindent
{\bf Equivariant down-sampling.}
Standard down-sampling operations are not equivariant as they impose a fixed sampling lattice on the feature space\cite{cohen2016group, wu2025measuring}. Let $F \in \mathbb{R}^{H\times W \times (n\times t)}$ be a feature map in a $\mathcal{G}$-equivariant network layer, and $\mathcal{G}=\{1,\cdots, t\}$. we have
\begin{equation}
    \label{eq:equi_feature_detail}
    F = \Big[\underbrace{F_1^{1}, \cdots, F_t^{1}}_{|\mathcal{G}|=t}, \cdots, \underbrace{F_1^{n}, \cdots, F_t^{n}}_{|\mathcal{G}|=t} \Big]
\end{equation}
Our method is to construct a $\mathcal{G}$-equivariant down-sampling operator by ensuring the sampling grid effectively co-transforms along with the features.
Firstly, we disentangle F into $t$ sub-maps $\{F_g\}_{g\in \mathcal{G}}$, where each $F_g \in \mathbb{R}^{H\times W\times n}$ consists of the all channels corresponding to the element $g\in\mathcal{G}$.
The key insight is to map each feature map $F_g$ back to a canonical one, then apply the standard down-sampling, and finally transform the result back to its original group index.

Formally, for any group element $g \in \mathcal{G}$, the corresponding output feature map $\tilde{F}_g$ is computed as:
\begin{equation}
	\label{eq:g_down-sampling}
	\tilde{F}_g = \left[\pi(g) \circ \text{Down}_s \circ \pi(g^{-1}) \right](F_g)  , \ \forall g \in \mathcal{G}.
\end{equation}
where $\pi(\cdot)$ denotes the group transformation operator, $\text{Down}_s$ is a $s$-stride down-sampling operator, and $\circ$ represents function composition. And all processed sub-maps $\{\tilde{F}_g\}_{g \in \mathcal{G}}$ are re-interleaved to construct the final output $F_{\text{out}} \in \mathbb{R}^{rH \times rW \times (n \times t)}$, like Eq.(\ref{eq:equi_feature_detail}) , which restores the standard $\mathcal{G}$-equivariant channel ordering. Specific examples and proofs of equivariance are provided in the supplementary materials.

\noindent
{\bf Equivariant Up-Sampling (Pixel-Shuffle).}
Standard up-sampling layers, such as Pixel-Shuffle~\cite{shi2016real}, are not group-equivariant by nature, as their fixed spatial reshaping pattern disrupts the channel-wise encoding of group transformations. To overcome this limitation, we introduce an Equivariant Pixel-Shuffle (E-PS) layer that preserves the group structure of features throughout the up-sampling process.

Let $F \in \mathbb{R}^{H\times W \times (r^2n\times t)}$ be the input feature map for an $r\times$ up-sampling. Our method also begins by disentangling $F$ into $t$ sub-maps $\{F_g\}_{g \in \mathcal{G}}$, where each $F_g \in \mathbb{R}^{H \times W \times (r^2n)}$.
To apply a standard up-sampling operation isotropically, we formulate a composite function that first maps each $F_g$ to a canonical one, applies Pixel-Shuffle, and then transforms the result back to its original one. This entire sequence yields the upsampled and correctly re-oriented sub-map $\tilde{F}_g$:
\begin{equation}
    \tilde{F}_g = \left[\pi(g) \circ \text{PS}_r \circ \pi(g^{-1})\right](F_g), \quad \forall g \in \mathcal{G}
    \label{eq:composite_transformation}
\end{equation}
where $\pi(\cdot)$ denotes the group transformation , $\text{PS}_r(\cdot)$ is the standard Pixel-Shuffle operator that reshapes a tensor of $r^2C$ channels to $C$ channels with $r \times$ larger spatial dimensions, and $\circ$ represents function composition.

Finally, all processed sub-maps $\{\tilde{F}_g\}_{g \in \mathcal{G}}$ are re-interleaved to construct the final output $F_{\text{out}} \in \mathbb{R}^{rH \times rW \times (n \times t)}$, like Eq.(\ref{eq:equi_feature_detail}) , which restores the standard $\mathcal{G}$-equivariant channel ordering. Specific examples and proofs of equivariance are provided in the supplementary materials.

\subsection{EQ-LayerNorm.}
A standard Layer Normalization (LN) layer includes a learnable affine transformation with scaling parameters $\bm{\gamma}$ and shifting parameters $\bm{\beta}$, both in $\mathbb{R}^{C}$. Applying this naively would assign an independent parameter to each channel for equivariant features $\bm{z} \in \mathbb{R}^{N \times C}$, thus breaking the structural dependency within each group-related block of channels and violating equivariance.

To enforce equivariance, we introduce a Group Equivariant Layer Normalization (EQ-LN) layer that employs parameter sharing. After the standard normalization step, we apply a shared affine transformation. Specifically, we define a reduced set of learnable parameters $\bm{\gamma}, \bm{\beta} \in \mathbb{R}^{C}$. Specifically, $\bm{\gamma}\ \text{and} \ \bm{\beta}$ have a weight-sharing along the group dimension. This ensures that all features belonging to the same group orbit are transformed by the exact same learnable parameters. Consequently, our EQ-LN layer commutes with the group action on the channel axis, rigorously preserving the network's group equivariance. More details can be found in the supplementary materials.

\subsection{Theoretical Analysis.}
In this section, we provide a rigorous theoretical analysis to establish the formal properties of our proposed architecture. Our analysis is twofold: first, we prove that our model achieves exact group equivariance to rotations and reflections (Theorem \ref{Thm1}). Second, we derive a novel generalization error bound (Theorem \ref{Thm2}) that quantifies the benefits of this equivariant design, formally linking the imposed symmetry to improved sample complexity.
\begin{Thm}\label{Thm1}
Let $\Phi_{eq}(\cdot)$ denotes an equivariant transformer including $L$-layer equivariant Self-Attention defined in Eq.(\ref{eq:equi_sa}). For an image $\mathbf{x}$ with size $H\times W\times c_0$, then the following result is satisfied for $\forall \tilde{g} \in S$:
\begin{equation}
    \Phi_{eq}\left[\pi_{\tilde{g}}\right] \left(\mathbf{x}\right) = \pi_{\tilde{g}} \left[\Phi_{eq}\right] (\mathbf{x}),
\end{equation}
where $\pi_{\tilde{g}}$ is a group transformation on the feature map and $\left[\cdot\right]$ denotes the composition of functions.
\end{Thm}

Theorem \ref{Thm1} shows that the proposed equivariant Vision Transformer is theoretically exact group equivariant with the rotation and mirror reflection. The proof can be found in the supplementary materials.

\begin{Thm}\label{Thm2}
For an image $\mathbf{x}$ with size $H\times W\times c_0$, and a equivariant transformer $\Phi_{eq}(\cdot)$ including $L$-layer equivariant self-attention, whose embedding dimension of the $i^{th}$ layer is $d_i$, for the group $S$, $|S|=t$, and activation function is set as GELU. If the latent continuous function of the $c^{th}$ channel of $\mathbf{x}$ denoted as $r_c: \mathbb{R}^2 \rightarrow \mathbb{R}$, the weight of patch embedding as $W_0$, and the $l^{th}$ linear layer weight of query, key, and value as $W^Q_{(l)}$, $W^K_{(l)}$, and $W^V_{(l)}$. The following conditions are satisfied:
    \begin{equation}
        \begin{split}
            & |r_c(x)| \leq F_0, \|W_0\|_{op} \leq \rho_0\\
            & \|W^Q_{(l)}\|_{op} \leq \rho_Q^{(l)}, \|W^K_{(l)}\|_{op} \leq \rho_K^{(l)}, \|W^V_{(l)}\|_{op} \leq \rho_V^{(l)}, \\
        \end{split}
    \end{equation}
    Then we have:
\begin{equation}\label{main_conclusion}
\begin{split}
    GE (\Phi_{eq}) \leq \frac{C}{\sqrt{n}} \frac{1}{t^{1/2}} + O(\sqrt{\frac{\operatorname{log}(1/\delta)}{n}}),
\end{split}
\end{equation}
where $C$ is determined by embedding dimensions $d$, the norm of the each linear layer, and the smoothness bounds of the input function. 
\end{Thm} 
Theorem~\ref{Thm2} derives the generalization error bound of our proposed Vision Transformer, establishing a formal upper bound under the action of the group Vision Transformers. This is essentially because isovariant networks introduce parameter sharing; fewer parameters mean lower model complexity, thus the network generalization error can be expected to be reduced. A precise derivation is provided in the supplementary materials.

%% file: sec/4_exp.tex
\section{Experiments}
In this section, we integrate the proposed equivariant modules into commond used ViT and Swin Transformer architectures, constructing strictly equivariant networks for image classification, object detection, and semantic segmentation. The modules are further extended to SwinIR for low-level vision tasks, including image super-resolution and video arbitrary-scale super-resolution.
\begin{table}[t]
\centering
\setlength{\tabcolsep}{5pt}
\renewcommand\arraystretch{1.0}
\caption{Classification performance comparison on ImageNet-1K and \textit{mini}ImageNet. All images are of size $224 \times 224$. T, S, and B denote the tiny, small, and base scales, respectively. And the suffixes -EQ and -EQ-R denote rotation equivariance and mirror reflection equivariance, respectively.}
\label{tab:table_imagenet1k}
\vspace{-4pt}
\begin{adjustbox}{width=0.9\columnwidth}
\begin{small}
    \begin{tabular}{l|c c c c}
    \Xhline{1.0pt}
    \multirow{2}{*}{Model} & Image & \multirow{2}{*}{Param.} & Top-1 & Top-5 \\
    & Size & & (\%) & (\%) \\
    \hline
    \multicolumn{5}{c}{\textbf{(a) ImageNet-1K}} \\
    \hline
    RegNetY-4G~\cite{radosavovic2020designing} & $224^2$ &  21M & 79.23 & 94.65 \\
    RegNetY-8G~\cite{radosavovic2020designing} & $224^2$ &  39M & 79.88 & 94.83 \\
    RegNetY-16G~\cite{radosavovic2020designing} & $224^2$ &  84M & 79.88 & 94.83 \\
    DeiT-T\cite{touvron2021training} & $224^2$ &  6M & 72.20 & 91.10 \\
    DeiT-S\cite{touvron2021training} & $224^2$ &  22M & 79.90 & 95.00 \\
    DeiT-B\cite{touvron2021training} & $224^2$ &  86M & 81.80 & 95.60 \\
    \hline
    ViT-T~\cite{dosovitskiy2020image} & $224^2$ & 6M & 74.28 & 91.86 \\
    \rowcolor{gray!15}
    ViT-T-EQ & $224^2$ & 8M & 74.41 & 91.88 \\
    \rowcolor{gray!15}
    ViT-T-EQ-R & $224^2$ & 7M & 74.57 & 91.90 \\
    ViT-S~\cite{dosovitskiy2020image} & $224^2$ & 22M & 78.23 & 93.87 \\
    \rowcolor{gray!15}
    ViT-S-EQ-R & $224^2$ & 17M & 78.69 & 94.24 \\
    ViT-B~\cite{dosovitskiy2020image} & $224^2$ & 87M & 77.18 & 93.11 \\
    \rowcolor{gray!15}
    ViT-B-EQ-R & $224^2$ & 38M & 80.28 & 94.91 \\
    \hline
    Swin-T & $224^2$ & 28M & 80.99 & 95.41 \\
    \rowcolor{gray!15}
    Swin-T-EQ & $224^2$ & 28M & 82.07 & 95.92 \\
    \rowcolor{gray!15}
    Swin-T-EQ-R & $224^2$ & 31M & 81.10 & 95.54 \\
    \Xhline{1.0pt}
    \multicolumn{5}{c}{\textbf{(b) \textit{mini}ImageNet}} \\
    \hline
    DeiT-S~\cite{touvron2021training} & $224^2$ & 22M &  70.83 & 89.74 \\
    DeiT-B~\cite{touvron2021training} & $224^2$ & 86M &  72.43 & 90.14 \\
    XCiT-S24~\cite{ali2021xcit} & $224^2$ & 26M & 85.79 & 96.31 \\
    XCiT-M24~\cite{ali2021xcit} & $224^2$ & 84M & 86.80 & 96.38 \\
    Swin-T~\cite{liu2021swin} & $224^2$ & 28M & 87.06 & 97.46 \\
    \rowcolor{gray!15}
    Swin-T-EQ-R & $224^2$ & 14M & 87.08 & 97.52 \\
    \Xhline{1.0pt}
    \end{tabular}
\end{small}
\end{adjustbox}
\vspace{-4mm}
\end{table}

\subsection{Image Classification}
{\bf Datasets and Training Details.} For image classification, the ImageNet-1K \cite{krizhevsky2012imagenet} dataset contains 1.28M training Images and 50K validation images with a total of 1,000 classes. \emph{mini}ImageNet includes 50K training images and 10K validation images across 100 categories. Specifically, images are trained and evaluated in $224 \times 224$ size for image classification. The top-1 and top-5 accuracy on the validation set is adopted as the evaluation metrics. For ViT, we adopt the republic github repository\footnote{https://github.com/huggingface/pytorch-image-models} to conduct all experiments. For Swin-Transformer, we follow the official release code to conduct all experiments. The training setting follows the image classification experiments are conducted on ImageNet-1K \cite{krizhevsky2012imagenet} and \emph{mini}ImageNet \cite{vinyals2016matching} datasets. In addition, We denote the variants of baseline models enhanced with rotation equivariance and reflection equivariance using the suffixes -EQ and -EQ-R, respectively.

\noindent
{\bf Quantitative Results.} As shown in Table~\ref{tab:table_imagenet1k}, our method demonstrates notable improvements across various Vision Transformer architectures on the ImageNet-1K and \textit{mini}ImageNet datasets. These quantitative results validate that our proposed equivariant methods effectively enhance classification accuracy. It is a well-documented observation that ViT, when trained from scratch without large-scale pre-training, often struggle to converge on smaller datasets. Consequently, we do not report results for ViT on the \textit{mini}ImageNet benchmark.

\begin{table*}[t]
\small
\centering 
\caption{Quantitative comparison (average PSNR/SSIM) for classical image SR on benchmark datasets.}
\label{tab:sr_results_reordered_params}
\vspace{-2mm}
\begin{tabular}{@{}lccrrrrrrrrrr@{}}
\toprule
\multirow{2.5}{*}{Method} & \multirow{2.5}{*}{Scale} & \multirow{2.5}{*}{Para.} & \multicolumn{2}{c}{Urban100} & \multicolumn{2}{c}{BSD100} & \multicolumn{2}{c}{Set14} & \multicolumn{2}{c}{Set5} & \multicolumn{2}{c}{Manga100} \\
\cmidrule(lr){4-5} \cmidrule(lr){6-7} \cmidrule(lr){8-9} \cmidrule(lr){10-11} \cmidrule(lr){12-13}
& & & PSNR & SSIM & PSNR & SSIM & PSNR & SSIM & PSNR & SSIM & PSNR & SSIM \\
\midrule
RCAN~\cite{zhang2018image} & $\times$2 & 15.4M & 33.34 & 0.9384 & 32.41 & 0.9027 & 34.12 & 0.9216 & 38.27 & 0.9614 & 39.44 & 0.9786 \\
SAN~\cite{dai2019second} & $\times$2 & 15.7M & 33.10 & 0.9370 & 32.42 & 0.9028 & 34.07 & 0.9213 & 38.31 & 0.9620 & 39.32 & 0.9792 \\
IGNN~\cite{zhou2020cross} & $\times$2 & - & 33.23 & 0.9383 & 32.41 & 0.9025 & 34.07 & 0.9217 & 38.24 & 0.9613 & 39.35 & 0.9786 \\
HAN~\cite{niu2020single} & $\times$2 & 16.1M & 33.35 & 0.9385 & 32.41 & 0.9027 & 34.16 & 0.9217 & 38.27 & 0.9614 & 39.46 & 0.9785 \\
NLSA~\cite{mei2021image} & $\times$2 & - & 33.42 & 0.9394 & 32.43 & 0.9027 & 34.08 & 0.9231 & 38.34 & 0.9618 & 39.59 & 0.9789 \\
SwinIR~\cite{liang2021swinir} & $\times$2 & 11.8M & 33.44 & 0.9399 & 32.45 & 0.9030 & \textbf{34.16} & \textbf{0.9232} & 38.32 & 0.9619 & 39.57 & \textbf{0.9789} \\
\rowcolor{gray!15}
SwinIR-EQ & $\times$2 & 5.2M & \textbf{33.54} & \textbf{0.9409} & \textbf{32.46} & \textbf{0.9032} & 34.10 & 0.9227 & \textbf{38.38} & \textbf{0.9620} & \textbf{39.59} & 0.9788 \\
\bottomrule
\vspace{-4mm}
\end{tabular}
\end{table*}

\noindent
{\bf Objection Detection and Semantic Segmentation.} 
For object detection, experiments are conducted on COCO 2017, which contains 118K training, 5K validation, and 20K test-dev images. For semantic segmentation, the ADE20K \cite{zhou2019semantic} dataset cover a broad range of 150 semantic categories. It's 20K for training, 2K for validation, and another 3K for testing. Using Swin-T as our visual encoder, we integrate it with Mask R-CNN~\cite{he2017mask} for detection and UPerNet~\cite{xiao2018unified} for segmentation. Compared to the baseline, our equivariant model demonstrates notable performance gains. On ADE20K, we improve the semantic segmentation mIoU from 44.51 to \textbf{44.86}. For object detection on COCO, our approach elevates the box mAP from 43.7 to \textbf{45.4} (+1.7) and the mask mAP from 39.8 to \textbf{41.2} (+1.4).

\subsection{Image Super-Resolution}
{\bf Datasets and Training Details.}
Following the setup in previous works \cite{liang2021swinir, zhang2022accurate},  we employ DIV2K~\cite{agustsson2017ntire} to train classic SR models.
Then we use Urban100~\cite{huang2015single}, BSD100~\cite{martin2001database}, Set14~\cite{zeyde2010single}, Set5~\cite{bevilacqua2012low}, and Manga109~\cite{matsui2017sketch} to evaluate the effectiveness of different SR methods. 
All comparison methods were trained in the same way as the original method, and the proposed method was trained using the same strategy as the baseline SwinIR~\cite{liang2021swinir}. In image super-resolution, we consider cases of rotational equivalence.

\begin{figure}[t]
    \centering
    \setlength{\abovecaptionskip}{0.1cm}
    \includegraphics[width=8.3cm]{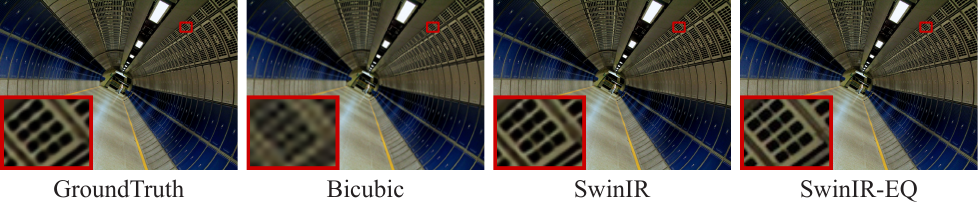}
    \vspace{-5mm}
    \caption{The visual comparison of the 4 times super resolution results from various methods on img078 of the Urban100.}
    \label{fig:swin}
    \vspace{-4mm}
\end{figure}

\noindent
\textbf{Quantitative Results.}
The results in Table~\ref{tab:sr_results_reordered_params} show that our proposed SwinIR-EQ achieves a compelling trade-off between performance and parameter efficiency. Compared to its powerful baseline, SwinIR-EQ demonstrates clear performance gains across most benchmark datasets for $\times$2 super-resolution. For instance, on the structurally rich Urban100 dataset, our method improves the PSNR for both scaling factors. Remarkably, these accuracy gains are achieved with a significantly more compact model. The SwinIR-EQ ($\times$2) utilizes only \textbf{5.2M} parameters, which is less than half the size of SwinIR (11.8M) and a small fraction of larger models like RCAN (15.4M). This dual improvement strongly validates our core hypothesis: by embedding rotation equivariance as an inductive bias, the model learns a more powerful and generalizable representation, reducing the need for a large parameter count while delivering superior restoration quality. The visual results are shown in Figure~\ref{fig:swin}.

Ablation experiments and further experimental results can be found in the supplementary materials.

\vspace{-2mm}

\subsection{Video Arbitrary Super-Resolution}
{\bf Datasets and Training Details.}All methods trained on the REDS dataset~\cite{nah2019ntire}, which comprises $240$ videos of resolution $720\times1,280$ captured by GoPro. Each video consists of $100$ HR frames. 
Following the settings in~\cite{chen2023motif, shang2024arbitrary}, we generate LR frames using the bicubic degradation model, with randomly sampled scaling factors from a uniform distribution $\mathcal{U}[1, 4]$. We test on the validation set of REDS comprising $30$ videos. We train to follow the original method. Here, we consider cases of rotational equivalence.

\begin{table}[!t]
\footnotesize
\centering
\setlength{\abovecaptionskip}{2pt} 
\setlength{\belowcaptionskip}{0pt}
\caption{Quantitative comparison with state-of-the-art methods on the REDS validation set (PSNR$\uparrow$ / SSIM$\uparrow$). The best results are highlighted in boldface.}
\label{table:reds} 
\begin{tabular}{lccc} 
\toprule
Method & $\times 2$ & $\times 3$ & $\times 4$ \\ 
\midrule
Bicubic & 31.51/0.911 & 26.82/0.788 & 24.92/0.713 \\ 
EDVR~\cite{wang2019edvr} & 36.03/0.961 & 32.59/0.904 & 30.24/0.853 \\
ArbSR~\cite{wang2021learning} & 34.48/0.942 & 30.51/0.862 & 28.38/0.799 \\
EQSR~\cite{wang2023deep} & 34.71/0.943 & 30.71/0.867 & 28.75/0.804 \\ 
RDN~\cite{zhang2018residual} + LTE~\cite{lee2022local} & 34.63/0.942 & 30.64/0.865 & 28.65/0.801 \\ 
\rowcolor{gray!15}
RDN-EQ + LTE-EQ & 34.73/0.943 & 30.68/0.867 & 28.68/0.802 \\ 
SwinIR~\cite{liang2021swinir} + LTE~\cite{lee2022local} & 34.73/0.943 & 30.73/0.866 & 28.75/0.804 \\ 
\rowcolor{gray!15}
SwinIR-EQ + LTE-EQ & 34.79/0.945 & 30.77/0.868 & 28.78/0.804 \\ 
\bottomrule
\vspace{-6mm}
\end{tabular}
\end{table}

\noindent
{\bf Quantitative Results.} To further demonstrate the versatility of our approach, we apply our method to the task of video arbitrary super-resolution. As shown in Table~\ref{table:reds}, our equivariant models consistently outperform their non-equivariant counterparts across all scaling factors ($\times 2, \times 3, \times 4$). These results clearly indicate that embedding equivariance is an effective strategy that can further enhance the quality of video restoration.

\vspace{-2mm}

%% file: sec/5_con.tex
\section{Conclusion}
In this work, we present a comprehensive framework to instill group equivariance into Vision Transformers by systematically redesigning each of its core components. Our approach constructs a holistic equivariant architecture, beginning with an equivariant patch embedding and extending to the self-attention mechanism through equivariant linear projections. Crucially, we introduce novel equivariant positional encodings by mapping absolute and relative coordinates to canonical representatives of their symmetry orbits, and preserve architectural integrity through equivariant resampling layers and group-aware normalization. This principled design culminates in a model that is not only provably equivariant, as supported by our theoretical analysis, but also empirically demonstrates consistent improvements in performance and data efficiency. The successful application of our framework to the Swin Transformer underscores its versatility and practical value, paving the way for a new class of Vision Transformers that possess a more fundamental and robust understanding of the visual world.